\documentclass[letterpaper,10pt,conference]{ieeeconf}
\IEEEoverridecommandlockouts
\usepackage{amsmath,amssymb,bm,booktabs,graphicx,microtype,cite}
\usepackage{tikz}
\usepackage{balance}
\usetikzlibrary{arrows.meta,positioning,calc,fit,backgrounds}
\usepackage[hidelinks]{hyperref}
\newcommand{\E}{\mathbb{E}}
\newcommand{\R}{\mathbb{R}}
\newcommand{\grad}{\nabla}

\title{\LARGE\bf Flow-Matched Motion Priors:\protect\\ Online Optimal-Transport Rewards for Imitation Learning} 

\author{Yilin Zou$^{1}$, Chenghua Liu$^{2}$, Chenglong Wu$^{1}$,
and Fanghua Jiang$^{1}$%
\thanks{This work was supported by the National Natural Science Foundation of
China (Grant Nos. 12472355 and 12532018).}%
\thanks{$^{1}$Yilin Zou and Chenglong Wu are Ph.D. candidates, and Fanghua
Jiang is a professor with the School of Aerospace Engineering, Tsinghua
University, Beijing 100084, China.
{\tt\small zouyl22@mails.tsinghua.edu.cn, jiangfh@tsinghua.edu.cn}.
Corresponding author: Fanghua Jiang.}%
\thanks{$^{2}$Chenghua Liu is a Ph.D. candidate with the Institute of
Software, Chinese Academy of Sciences, Beijing, China.
{\tt\small liuch.russell@gmail.com}.}%
}

\hypersetup{
  pdftitle={Flow-Matched Motion Priors: Online Optimal-Transport Rewards for Imitation Learning},
  pdfauthor={Yilin Zou, Chenghua Liu, Chenglong Wu, Fanghua Jiang}
}
\begin{document}
\maketitle

\begin{abstract}
Learning a motion prior requires a reward that guides a policy from its current
behavior toward demonstrated motion. Adversarial Motion Priors (AMP) provide
such a reward with a discriminator. However, adversarial objectives can become
uninformative when policy and expert supports are far apart. A naive use of
optimal transport (OT) averages matched expert successors into a barycentric
target. Averaging across gait phases can weaken the target's joint motion.
We introduce Flow-Matched Motion Priors (FMP), an online scalar
reward learned from paths connecting current rollout histories to an expert
motion bank. Entropic OT supplies the coupling. Before each policy update,
we train a neural potential with flow matching (FM) along the rollout-to-expert
paths, endpoint-gradient supervision, and relative-value calibration. The actor receives only
physical observations and the reward remains a scalar, as in AMP. Controlled
reward-model experiments show substantially better generalization beyond the
fitting rollout than value-only or endpoint-only fitting. On Unitree G1, matched
50-million-transition experiments compare FMP with AMP, a barycentric OT reward,
and nested ablations under demonstration and fixed-pose initialization. FMP
produces stable forward walking at 0.727\,m/s from demonstration resets and
0.338\,m/s from a fixed default pose. In the fixed-pose condition, it incurs
129 falls versus 243 for the endpoint-only control.
Against a static score-gradient teacher, dynamic FM
reduces score-increment error at interpolation fractions 0.25 and 0.50 and
uses 29\% less offline fitting time.
\end{abstract}

\section{Introduction}
Motion priors turn demonstrations into rewards for learning physically
simulated skills. Their appeal is that a policy can acquire characteristic
movement without tracking a prescribed reference phase at deployment.
Adversarial Motion Priors (AMP) learn such a reward by discriminating between
policy and expert motion~\cite{peng2021amp}. As in other adversarial imitation
objectives, a Jensen--Shannon (JS)-style comparison can provide little useful
signal when the two supports are far apart~\cite{ho2016gail}. This motivates a
geometric reward whose supervision remains defined between the distributions.

Optimal transport (OT) compares distributions through a coupling, with the
Earth Mover's Distance (EMD) as a classical instance. Transport couplings
have already been used to construct imitation rewards
~\cite{xiao2019wail,dadashi2021pwil}. Entropic Sinkhorn computation makes
minibatch couplings practical~\cite{cuturi2013sinkhorn}. The most direct use
of this coupling is a \emph{naive OT} reward: average the matched expert
successors and penalize the distance to that barycentric target. A key
difficulty is that averaging successors from different gait phases can weaken
joint motion in the target. This can happen even when individual expert samples
contain strong joint motion. The barycentric target can therefore lose motion
signals present in the demonstrations. FMP keeps the OT coupling and
explicitly transports from the current rollout to the expert bank. The sampled
rollout-to-expert displacements provide inexpensive supervision for a neural
scalar potential through flow matching (FM). A policy self-transport term
corrects the policy-to-expert score. The potential is trained online and
supplies a reward to proximal policy optimization (PPO) through the same
scalar interface used by AMP~\cite{schulman2017ppo}. This construction draws
on established links between flow supervision and scalar potentials
~\cite{liu2022rectifiedot,neklyudov2023action,balcerak2025energy}.
We apply these ideas to an online motion prior.

Turning path supervision into a stable PPO reward imposes two practical
requirements. First, matching gradients does not determine relative scalar
values in sparsely sampled regions. We therefore calibrate the scalar values
used by PPO. Second, the reward model must track the policy as its rollout
distribution changes. We therefore refresh the rollout-to-expert transport
supervision after every rollout. We fit the reward before estimating policy
returns and average model parameters only within these fitting updates.
The actor itself receives only physical observations: it observes neither a
generated target nor a latent motion condition.

The paper makes three contributions. First, it introduces a scalar motion
reward learned by FM along rollout-to-expert paths. The reward combines value
calibration with a policy self-transport correction. Second, it tests the role
of path supervision. This supervision improves reward predictions on motion
histories beyond the fitting rollout. Compared with static supervision,
dynamic FM has lower increment error at interpolation fractions 0.25 and 0.50
and takes less time to fit. Static supervision is more accurate near the
expert endpoint. Third, it evaluates humanoid skill acquisition from scratch
under two initialization regimes. All methods use identical policy
architectures and equal interaction budgets. Motion metrics distinguish
walking from standing. Fixed-pose initialization deliberately places the
initial policy far from the demonstration support and provides the main
physical test.

\begin{figure*}[t]
\centering
\resizebox{.98\textwidth}{!}{
\begin{tikzpicture}[
  x=1cm,y=1cm,>=Stealth,
  font=\sffamily\fontsize{7}{8}\selectfont,
  line cap=round,line join=round,
  arrow/.style={->,draw=black!60,line width=.65pt},
  annotation/.style={font=\sffamily\fontsize{6.3}{7.2}\selectfont,text=black!65},
  heading/.style={font=\sffamily\fontsize{8}{9}\selectfont\bfseries,text=black!85},
  softpair/.style={draw=tpblue!35,densely dashed,line width=.4pt},
  policypt/.style={circle,draw=white,line width=.4pt,fill=tpblue,
    inner sep=0pt,minimum size=3.3pt},
  expertpt/.style={circle,draw=white,line width=.4pt,fill=tpgreen,
    inner sep=0pt,minimum size=3.3pt}
]
\definecolor{tpblue}{HTML}{2466AC}
\definecolor{tpgreen}{HTML}{288461}
\definecolor{tpamber}{HTML}{D78021}
\definecolor{tpink}{HTML}{354255}
\path[use as bounding box] (0,-.93) rectangle (14.45,3.90);

\node[heading] at (1.98,3.66) {1\quad Match current motion};
\node[heading] at (6.75,3.66) {2\quad Fit flow-matched potentials};
\node[heading] at (11.83,3.66) {3\quad Score realized motion};
\node[annotation,text=black!45,anchor=east] at (14.30,3.29) {schematic feature geometry};

\path[fill=tpblue!5] (1.08,1.83) ellipse [x radius=.78,y radius=.60];
\path[fill=tpgreen!6] (2.98,2.46) ellipse [x radius=.66,y radius=.55];
\coordinate (p1) at (.54,1.77);
\coordinate (p2) at (.82,2.13);
\coordinate (p3) at (.94,1.72);
\coordinate (p4) at (1.29,2.06);
\coordinate (p5) at (1.41,1.52);
\coordinate (p6) at (.89,1.40);
\coordinate (p7) at (1.62,1.85);
\coordinate (p8) at (1.13,2.30);
\coordinate (e1) at (2.46,2.29);
\coordinate (e2) at (2.71,2.65);
\coordinate (e3) at (3.06,2.59);
\coordinate (e4) at (3.32,2.85);
\coordinate (e5) at (3.43,2.43);
\coordinate (e6) at (3.16,2.08);
\coordinate (e7) at (2.75,2.17);
\coordinate (e8) at (2.98,2.91);
\foreach \src/\dst in {p1/e2,p1/e6,p2/e3,p2/e8,p3/e1,p3/e5,
  p4/e2,p4/e3,p4/e4,p5/e5,p5/e6,p6/e7,p7/e3,p7/e6,p8/e4,p8/e8}
  \draw[softpair] (\src) -- (\dst);
\draw[tpblue!70,line width=.75pt,->] (p4) -- (e3);
\draw[tpblue!65,line width=.55pt,dashed,->] (p1) to[bend right=35] (p5);
\draw[tpblue!65,line width=.55pt,dashed,->] (p5) to[bend right=22] (p3);
\foreach \p in {p1,p2,p3,p4,p5,p6,p7,p8} \node[policypt] at (\p) {};
\foreach \p in {e1,e2,e3,e4,e5,e6,e7,e8} \node[expertpt] at (\p) {};
\node[annotation,text=tpblue] at (.97,2.66) {current rollout};
\node[annotation,text=tpgreen] at (2.99,3.17) {fixed expert bank};
\node[annotation,text=tpblue!85!black] at (.83,1.12) {policy self-OT};
\node[annotation] at (2.56,1.40) {soft cross-OT};
\node[annotation,text=black!70] at (2.03,.82) {Current source banks refresh online};
\draw[arrow] (3.79,2.14) -- (4.36,2.14);
\node[annotation,align=center] at (4.07,2.52) {sample\\a pair};

\coordinate (source) at (4.82,1.32);
\coordinate (target) at (8.51,2.83);
\coordinate (query) at (6.665,2.075);
\foreach \x/\y in {4.91/2.02,5.52/2.30,6.16/1.42,7.19/1.63,7.46/2.65,8.02/2.09}
  \draw[draw=tpblue!30,line width=.6pt,->] (\x,\y) -- ++(.45,.184);
\draw[draw=tpblue!35,line width=.8pt] (source) -- (target);
\draw[draw=tpblue,line width=1.05pt,->] (source) -- ($(source)!.34!(target)$);
\draw[draw=tpblue,line width=1.05pt,->] ($(source)!.40!(target)$) -- ($(source)!.69!(target)$);
\draw[draw=tpblue,line width=1.05pt,->] ($(source)!.75!(target)$) -- (target);
\node[policypt,minimum size=4.6pt] at (source) {};
\node[expertpt,minimum size=4.6pt] at (target) {};
\node[circle,fill=tpamber,draw=white,line width=.5pt,minimum size=4.6pt,inner sep=0pt] at (query) {};
\draw[draw=tpamber!80!black,line width=.45pt] ($(query)+(0,.09)$) -- (6.665,2.63);
\node[annotation,text=tpamber!85!black] at (6.665,2.82) {interior training query};
\node[annotation,anchor=east] at (4.80,1.08) {flow time 0};
\node[annotation,anchor=south] at (8.44,3.02) {flow time 1};
\node[annotation,text=tpblue!85!black] at (7.15,1.29) {potential-gradient field};
\node[annotation,align=center] at (6.80,.82)
  {Value calibration + endpoint gradients\\+ straight-path supervision};
\draw[arrow] (9.02,2.14) -- (9.52,2.14);

\begin{scope}[shift={(11.89,2.05)},rotate=13]
  \path[fill=tpblue!4] plot[smooth cycle,tension=.78] coordinates
    {(-1.76,-.30) (-1.52,.54) (-.43,.93) (.88,.79) (1.72,.11) (1.21,-.64) (-.45,-.76)};
  \foreach \factor/\tone in {1/35,.77/45,.54/55,.31/65} {
    \begin{scope}[scale=\factor]
      \draw[draw=tpblue!\tone,line width=.6pt] plot[smooth cycle,tension=.78] coordinates
        {(-1.76,-.30) (-1.52,.54) (-.43,.93) (.88,.79) (1.72,.11) (1.21,-.64) (-.45,-.76)};
    \end{scope}
  }
\end{scope}
\foreach \x/\y in {11.22/2.67,12.72/2.16,12.18/1.63}
  \fill[tpink!45] (\x,\y) circle[radius=1.25pt];
\coordinate (observed) at (10.88,1.79);
\filldraw[fill=tpink,draw=white,line width=.55pt] (observed) circle[radius=2.05pt];
\node[annotation,text=tpink,anchor=east] at (10.63,1.47) {observed motion};
\node[annotation,text=tpblue!85!black] at (12.54,2.92) {learned score contours};
\node[annotation,text=black!80,align=center] at (11.87,.91)
  {Cross potential minus the mean\\of both policy-self potentials};

\draw[black!12,line width=.5pt] (.22,.57) -- (14.25,.57);

\foreach \a in {-.13,.11} {
  \foreach \b in {-.20,-.01,.18} {
    \draw[tpblue!35,line width=.38pt] (.73,\a) -- (1.08,\b);
    \draw[tpblue!35,line width=.38pt] (1.08,\b) -- (1.43,\a);
  }
}
\foreach \x/\y in {.73/-.13,.73/.11,1.08/-.20,1.08/-.01,1.08/.18,1.43/-.13,1.43/.11}
  \filldraw[fill=white,draw=tpblue,line width=.6pt] (\x,\y) circle[radius=1.4pt];
\node[annotation,text=black!85] at (1.08,-.44) {Physical-state policy};

\draw[tpink,line width=.9pt] (3.65,.16) circle[radius=.068];
\draw[tpink,line width=1pt] (3.65,.09) -- (3.65,-.11);
\draw[tpink,line width=.85pt] (3.45,-.02) -- (3.65,.035) -- (3.84,-.075);
\draw[tpink,line width=.85pt] (3.65,-.11) -- (3.49,-.24) -- (3.36,-.27);
\draw[tpink,line width=.85pt] (3.65,-.11) -- (3.79,-.20) -- (3.82,-.31);
\draw[black!25,line width=.4pt] (3.25,-.33) -- (3.99,-.33);
\node[annotation,text=black!85] at (3.65,-.44) {G1 simulation};

\foreach \x/\height in {6.10/.12,6.35/.25,6.60/.17} {
  \draw[draw=tpink!55,fill=white,line width=.55pt] (\x-.10,-.25) rectangle (\x+.10,.25);
  \draw[draw=tpblue,line width=1pt] (\x,-.15) -- (\x,\height);
}
\node[annotation,text=black!85] at (6.35,-.44) {Observed history};

\draw[black!25,line width=.4pt] (9.31,-.25) -- (9.31,.24);
\draw[black!25,line width=.4pt] (9.29,-.23) -- (9.93,-.23);
\draw[tpblue,line width=1pt] (9.32,-.19) .. controls (9.61,-.19) and (9.61,.16) .. (9.90,.16);
\node[annotation,text=black!85] at (9.60,-.44) {Scalar reward};

\draw[arrow,draw=tpink!75] (12.58,.20) arc[start angle=155,end angle=-145,radius=.22];
\node[font=\sffamily\fontsize{6.2}{7}\selectfont,text=tpink] at (12.78,.07) {PPO};
\node[annotation,text=black!85] at (12.78,-.44) {Policy optimization};

\draw[arrow] (1.65,-.01) -- (3.14,-.01);
\node[annotation,fill=white,inner sep=1pt] at (2.37,.13) {action};
\draw[arrow] (4.10,-.01) -- (5.93,-.01);
\draw[arrow] (6.86,-.01) -- (9.11,-.01);
\draw[arrow] (10.10,-.01) -- (12.20,-.01);
\node[annotation,fill=white,inner sep=1pt] at (11.13,.14) {returns / advantages};
\draw[arrow] (12.78,-.58) -- (12.78,-.80) -- (1.08,-.80) -- (1.08,-.58);
\node[annotation,fill=white,inner sep=1.2pt] at (6.8,-.80) {update policy; actor receives physical observations only};

\draw[arrow,draw=black!40,line width=.5pt] (3.65,.30) -- (3.65,.44) -- (1.08,.44) -- (1.08,.27);
\node[annotation,fill=white,inner sep=.8pt] at (2.37,.44) {physical observation};
\draw[arrow,draw=tpblue!65] (11.87,.64) -- (11.87,.36) -- (9.60,.36) -- (9.60,.26);
\draw[arrow,draw=tpink!55,densely dashed,line width=.5pt] (10.28,1.20) -- (observed);
\end{tikzpicture}}
\caption{FMP transports current rollout histories toward demonstrations and
fits a shared scalar potential on the resulting paths. The actor receives only
physical observations and the fitted scalar reward.}
\label{fig:overview}
\end{figure*}
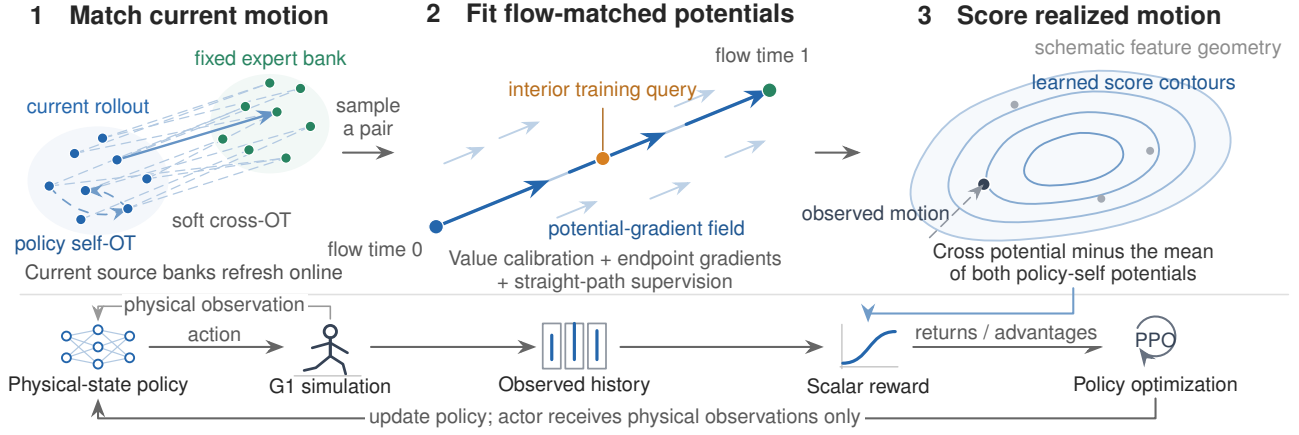

\section{Related Work}
\subsection{Motion priors and imitation rewards}
DeepMimic demonstrates reinforcement learning from reference motion through
tracking rewards~\cite{peng2018deepmimic}. AMP replaces explicit tracking with
a learned discriminator~\cite{peng2021amp}. Adversarial Skill Embeddings extend
this idea to controllable latent skills~\cite{peng2022ase}. Generative
Adversarial Imitation Learning establishes the corresponding occupancy-matching
view~\cite{ho2016gail}. Score-Matching Motion Priors (SMP) learn a reusable
diffusion prior and obtain rewards from its score-matching
objective~\cite{mu2026smp}. SMP illustrates how a pretrained generative motion
model can supply a scalar reward.
Our model is fitted online using transport between the current policy and
demonstrations. These methods inform our design. We use AMP as an
experimental baseline.

\subsection{Transport for imitation learning}
Wasserstein objectives offer a geometric alternative to adversarial distribution
comparison~\cite{arjovsky2017wgan,peyre2019computational}. Wasserstein
Adversarial Imitation Learning (WAIL) uses a learned transport potential as
an imitation reward~\cite{xiao2019wail}. Primal Wasserstein Imitation
Learning (PWIL) constructs rewards directly from primal policy--expert
matching~\cite{dadashi2021pwil}. Both methods use transport to provide a scalar
reward. They do not fit a time-indexed field along rollout-to-expert paths or
supervise a neural potential at intermediate points on those paths.
FMP retains the scalar reward interface and adds this path supervision,
endpoint derivatives, and scalar calibration. Entropic Sinkhorn computation
makes minibatch couplings practical~\cite{cuturi2013sinkhorn}.
Sinkhorn divergences remove self-bias~\cite{feydy2019sinkhorn}, and minibatch
averaging reduces variation in the empirical teacher~\cite{fatras2020minibatch}.
These results motivate our use of a debiased transport score and source-bank
resampling.

\subsection{Flow supervision and scalar potentials}
Flow matching regresses the velocity of a probability path without simulating
that path during training~\cite{lipman2023flow}. Rectified Flow and
OT-based conditional FM relate the coupling of endpoints to the resulting
vector field~\cite{liu2023rectified,tong2024cfm}. Gradient-field constraints
appear in cost-specific rectified flow with quadratic cost
~\cite{liu2022rectifiedot} and Action Matching~\cite{neklyudov2023action}.
Score matching provides a separate
precedent for learning a scalar model through its
derivatives~\cite{hyvarinen2005score}. More directly, Energy Matching combines
transport-gradient regression with energy-based
modeling~\cite{balcerak2025energy}. Projected Energy Matching studies the
relationship between unrestricted flows and conservative potential
models~\cite{barco2026projected}. These connections motivate our
parameterization. Our evaluation tests how these connections translate into
reward generalization and physical skill acquisition.

\section{Method}
\subsection{Motion representation and reward interface}
Let $s_t$ denote the physical observation at simulator step $t$, $a_t$ the
action, and $\pi_\phi(a_t\mid s_t)$ the policy with parameters $\phi$.
Let $g_t$ denote the per-frame motion feature, containing pose and velocity
information. The history $h_t=(g_{t-H+1},\ldots,g_t)$ has $H$ frames, and
$x_t=f(h_t)\in\R^d$ is its normalized representation under a fixed feature
map $f$. Expert histories use the same map and are denoted $y_j\in\R^d$,
where $j$ indexes an expert sample. The actor observation and motion-prior
history have different roles: the former determines the action; the latter
is scored after the action has been executed.

The barycentric baseline uses an OT coupling to construct a target for the
next frame. It conditions matching
probabilities $p_j(h_t)$ on the $H-1$ most recent frames of $h_t$. Let
$d_g$ denote the single-frame feature dimension. Write the normalized policy
successor as $u_{t+1}\in\R^{d_g}$ and the normalized expert successors as $y_j^+$.
For reward temperature $T>0$, the target and reward are
\begin{equation}
 \begin{aligned}
 \bar y^+(h_t)&=\sum_j p_j(h_t)y_j^+,\\
 r_t^{\rm bary}&=2\exp\!\left[-\frac{\|u_{t+1}-\bar y^+(h_t)\|^2}{d_g T}\right].
 \end{aligned}
 \label{eq:bary}
\end{equation}
Matching uses the preceding history, and the reward scores the new frame.
Our reward evaluates a learned
scalar on $x_{t+1}$, the realized successor history. It requires no target
to be selected or observed by the actor.

\subsection{A transport reference score}
Let $Q$ be the empirical pool of current rollout features, and let
$B=\{x_i\}_{i=1}^n$ be a uniformly sampled bank of $n$ features from $Q$.
The expert bank is $E=\{y_j\}_{j=1}^m$, containing $m$ features. The transport
direction used throughout is from the current rollout bank to this expert bank.
We solve entropically regularized OT between $B$ and $E$, and between $B$ and
itself, using quadratic cost $c(x,y)=\tfrac12\|x-y\|^2$. The entropy
coefficient is $\varepsilon_B>0$, and bank atoms have uniform positive
masses. With target duals $b_j$ and masses $w_j=1/m$, the negative dual
extension is
\begin{equation}
 A_E^B(x)=\varepsilon_B\log\sum_{j=1}^m w_j
 \exp\!\left[\frac{b_j-c(x,y_j)}{\varepsilon_B}\right].
 \label{eq:extension}
\end{equation}
The superscript $B$ records dependence on the sampled policy bank.
Let $p_j^B(x)$ denote the normalized terms inside the sum.
The operator $\grad$ differentiates with respect to the candidate feature $x$.
Then
\begin{equation}
 \begin{aligned}
 p_j^B(x)&=\frac{w_j e^{(b_j-c(x,y_j))/\varepsilon_B}}
 {\sum_l w_l e^{(b_l-c(x,y_l))/\varepsilon_B}},\\
 \grad A_E^B(x)&=\sum_jp_j^B(x)y_j-x.
 \end{aligned}
 \label{eq:gradient}
\end{equation}
Analogous extensions $A_{P,1}^B$ and $A_{P,2}^B$ use the two duals of
policy self-transport, with $P$ denoting the policy distribution.
The reference score, for which higher values are preferred, is
\begin{equation}
 S_B(x)=A_E^B(x)-\tfrac12\big[A_{P,1}^B(x)+A_{P,2}^B(x)\big].
 \label{eq:score}
\end{equation}
Keeping both self terms matches the numerical solver. At fixed entropy
coefficient, this score is the negative source first variation of the
corresponding debiased transport objective, up to an additive constant.
This standard identity motivates the sign and source correction~\cite{feydy2019sinkhorn}.

Source-bank resampling changes $S_B$ even when the policy is fixed. We
therefore fit the mean empirical potential
\begin{equation}
 \begin{aligned}
 \bar S_Q(x)&=\E_{B\sim Q}S_B(x),\\
 \grad\bar S_Q(x)&=\E_{B\sim Q}\grad S_B(x).
 \end{aligned}
 \label{eq:mean}
\end{equation}
The expectation averages over finite sampled banks. It need not equal the
population Sinkhorn potential. Each bank uses
$\varepsilon_B=\max\{0.1\,\mathrm{median}_{i,j}c(x_i,y_j),10^{-3}/d\}$.
We hold this value constant when differentiating the extensions. Fresh source banks
are drawn at every auxiliary update while the expert bank stays fixed.

\subsection{Why a scalar score differs from a fixed target}
Equation~\eqref{eq:gradient} still contains a conditional mean. Unlike the
fixed target in \eqref{eq:bary}, its matching weights change with the candidate
being scored. Let $C_p(x)$ be the covariance of
expert atoms under $p_j^B(x)$, and $I$ the identity matrix. Then
\begin{equation}
 \grad^2 A_E^B(x)=-I+C_p(x)/\varepsilon_B.
 \label{eq:curvature}
\end{equation}
For two equal-weight one-dimensional atoms at $-a$ and $a$, with $a>0$
and equal duals, the midpoint has curvature $-1+a^2/\varepsilon_B$.
It is a local minimum if $\varepsilon_B<a^2$. A reward based on proximity to
the fixed barycenter instead has its maximum at that midpoint. Thus averaging in a score
gradient need not impose a reward maximum at the averaged motion.
This observation concerns the cross-potential, not a guarantee about the
self-corrected learned score or PPO. An expected exponential reward over
sampled targets can also retain multiple modes. Target sampling and
policy-gradient averaging therefore do not necessarily cause collapse.

\subsection{Flow-matched, value-calibrated potentials}
The neural model has parameters $\theta$ and three scalar heads
$U_\theta^k(x,\tau)$ indexed by $k\in\{E,P1,P2\}$, corresponding to the
expert and two policy-self extensions. Flow time $\tau\in[0,1]$ is distinct
from simulator step $t$. The scalar score used for reward evaluation is
\begin{equation}
 \widehat S_\theta(x)=U_\theta^E(x,0)
 -\tfrac12\big[U_\theta^{P1}(x,0)+U_\theta^{P2}(x,0)\big].
 \label{eq:neural}
\end{equation}
All heads share a feature trunk. We parameterize each potential as a
quadratic base $-\|x\|^2/2$, a learned linear term, and a neural residual.
The common quadratic term cancels in \eqref{eq:neural}.

\emph{Scalar calibration.}
For a query sampled from the current fitting pool, define the scalar error
$\delta_B(x)=\widehat S_\theta(x)-S_B(x)$. With positive scale $c_0=0.10$,
the value loss is
\begin{equation}
 L_{\rm val}=\E_B\frac{\E_x[(\delta_B(x)-\E_x\delta_B(x))^2]}{c_0^2}.
 \label{eq:value}
\end{equation}
The inner expectation is estimated on the fitting minibatch. Centering
removes the arbitrary additive constant. This term directly supervises the
relative scalar values that become policy rewards. Agreement of gradients
at isolated observations does not determine those values between disconnected
or sparsely covered regions.

\emph{Endpoint supervision.}
The gradient of the reward score is fitted explicitly at flow time zero:
\begin{equation}
 L_{\rm end}=\E_{B,x}\big\|\grad\widehat S_\theta(x)-\grad S_B(x)\big\|^2.
 \label{eq:endpoint}
\end{equation}
Sampling a continuous flow time alone gives zero probability of selecting
exactly this endpoint. This separate loss therefore connects geometric
supervision to the function used for reward evaluation.

\emph{Interior path supervision.}
For head $k$, sample a target $z$ from the atoms of its cross- or self-transport
problem using its conditional probabilities at source query $x$.
Draw $\tau$ uniformly on $[0,1]$, and form the interpolation
$x_\tau=(1-\tau)x+\tau z$. With heads sampled uniformly, define
\begin{equation}
 L_{\rm flow}=\E_{B,x,k,z,\tau}
 \big\|\grad U_\theta^k(x_\tau,\tau)-(z-x)\big\|^2.
 \label{eq:flow}
\end{equation}
The complete loss is $L=L_{\rm val}+L_{\rm end}+L_{\rm flow}$, with
unit weights on these three terms. These interpolants are feature-space
training queries, not reference trajectories imposed on the robot.
The conditional extension from bank atoms to arbitrary rollout queries need
not preserve the expert marginal exactly. Moreover, the unrestricted
conditional velocity of a finite entropic coupling need not be conservative.
We use the scalar-gradient model as a constrained approximation and test its
resulting scores directly, without claiming exact distribution transport.

\subsection{Online fitting and policy optimization}
For each rollout, the policy is held fixed during 128 reward-model updates.
Histories from environments 0--767 supply neural fitting queries. All 1,024
environments contribute to the transport pool and receive fitted rewards.
Each update resamples a source bank, solves its cross- and self-transport
problems, and minimizes $L$. The trainable reward model and optimizer persist
across rollouts. To reduce fitting noise, an exponential moving average
(EMA) of parameters is formed within these 128 updates, with a half-life of
16 updates. At each rollout, the EMA restarts from the current trainable model.
This averaging reduces within-rollout variation without introducing an
additional average over older policies.

The averaged model scores all recorded successor histories. Let $m_\theta$
and $q_\theta$ be the median and interquartile range (IQR) of these scores.
We use a positive numerical floor $\eta$ and the logistic function
$\sigma(u)=1/(1+e^{-u})$. The reward delivered to PPO is
\begin{equation}
 r_t=2\sigma\!\left(
 \frac{\widehat S_\theta(x_{t+1})-m_\theta}{\max(q_\theta,\eta)}\right).
 \label{eq:reward}
\end{equation}
Returns and generalized advantage estimation (GAE)~\cite{schulman2016gae}
are computed from this reward before the policy update. The reward network
is then held fixed for that PPO update. Rollout normalization and bounding
make \eqref{eq:reward} a practical surrogate. Its policy gradient is not
the exact gradient of a Sinkhorn divergence. At deployment,
the actor runs without transport solves or the reward network.

\section{Experimental Setup}
\subsection{Robot, demonstrations, and training protocol}
We use Unitree G1 in Isaac Lab~\cite{mittal2025isaaclab} with the native
MimicKit PPO implementation~\cite{peng2025mimickit}. The demonstration is a
clean forward segment of LaFAN1 \texttt{walk1\_subject1}, source frames
2370--2580~\cite{harvey2020robust}. We retarget it to G1 and resample it from
30\,Hz to 50\,Hz. Simulation control also runs at 50\,Hz. Motion-prior inputs
concatenate ten 239-dimensional frames ($H=10$, $d=2{,}390$). All scalar
controls share this representation. The fixed feature map standardizes using
expert data, clips each coordinate to $[-10,10]$, and scales by $1/\sqrt d$.
The actor receives 237 physical observation values and no reference phase.
There is no additional forward-speed task reward.

We train separate runs with two reset regimes. \emph{Dataset reset}
samples a demonstration state and its associated history. \emph{Fixed-pose
reset} starts from the robot's default joint configuration, zero root speed,
and a history repeating the actual initial state. This deliberately creates a
large initial support gap from the demonstrations. Falls trigger an immediate
reset under the same regime. Evaluation uses the corresponding training
reset regime, including the repeated initial history for fixed-pose runs.

Each rollout contains 32 steps in 1,024 parallel environments, or 32,768
transitions. Episodes are capped at 500 steps (10\,s). Every method starts
from the same untrained actor and critic, and learned rewards start fresh.
The final comparison is at 50,003,968 transitions, the first complete rollout
at or above the prespecified 50-million budget. We retain evaluations at
5, 10, 20, 30, and 40 million transitions. All runs restart simulation
at the same continuation boundaries while restoring their learning states.
One training seed is used throughout. 

The potential trunk contains two 256-unit layers with sigmoid linear unit
(SiLU) activations and three scalar outputs. Time is encoded by
$(\tau,\tau^2,\sin\pi\tau,\cos\pi\tau)$. The nonlinear trunk receives
$\sqrt d\,x$ and this encoding. The base and linear terms use $x$.
Each fitting minibatch and transport bank contains 256 samples. The fixed expert bank is shared across
scalar controls. Adam optimizes reward parameters at $3\times10^{-4}$ with
gradient-norm clipping at 10. Policy and critic use their native stochastic
gradient descent optimizers at $10^{-4}$. PPO clipping is 0.2, discount is
0.99, and GAE decay is 0.95. Auxiliary optimizer counts are matched across
learned controls. We also report wall time because derivative losses change
the cost of each update. Effective training settings,
continuation boundaries, and provenance records accompany the supplementary results.

\subsection{Comparators and nested ablations}
\emph{FM potential} uses all three losses in \eqref{eq:value}--\eqref{eq:flow}.
\emph{Endpoint} removes only $L_{\rm flow}$. \emph{Value} additionally
removes $L_{\rm end}$. These are the central nested ablations: they share
architecture, initialization, feature map, scalar target, update budget,
and within-rollout averaging. \emph{16-bank score} evaluates the average
of 16 empirical transport scores directly, with the same reward normalization.
It tests the effect of approximating the scalar with a neural model.

\emph{AMP} uses the official MimicKit discriminator, binary cross-entropy,
replay, gradient penalty, and capped reward. Its feature normalization evolves
and its rollout is scored before updating the discriminator. The transport
methods use fixed feature normalization and fit the reward before scoring.
\emph{Barycentric OT} conditions matching on the preceding $H-1$ frames and
rewards proximity of only the latest frame to the expert-successor mean.
It retains fresh native expert banks and has no source-self correction.
Consequently, it tests the practical target construction, not the isolated
contribution of $L_{\rm flow}$.

A separate \emph{static-path} control tests the cost-quality tradeoff behind
the path construction. It retains the value and endpoint losses and uses the
same interpolation queries. An independent source bank $B'$ supplies a fresh
label for the scalar reward score:
\begin{equation}
 L_{\rm static}=\E\big\|\grad\widehat S_\theta(x_\tau)
                  -\grad S_{B'}(x_\tau)\big\|^2.
 \label{eq:static}
\end{equation}
This replaces the individual-head dynamic field with a static contrast
gradient and fixes neural time to zero. Dynamic and static models share
initialization, sampling streams, and 9,216 offline updates. Static labels
require an additional transport solve, which is included in fitting time.
This paired comparison measures the accuracy and cost of path supervision.
The 16-bank score separately tests direct scalar reward evaluation.

\subsection{Evaluation protocol and metrics}
Frozen deterministic policies are evaluated on 1,024 matched first episodes,
with detailed trajectories for the first 64 environments. We measure mean
episode length, fall count, and signed forward speed over valid observations
between 5 and 10\,s. Positive speed denotes progress along the demonstration
heading at reset. Trajectories that end before the late window are excluded
from the speed estimate, rather than assigned zero speed. Table~\ref{tab:final50m}
reports the number of contributing trajectories as $n$. Full-window coverage
is recorded in the supplementary results.

Joint excursion uses the six sagittal leg joints: bilateral hip pitch, knee,
and ankle pitch. In each complete one-second window, joint range is the
95th minus 5th percentile. Excursion averages these ranges relative to the
expert after excluding the first second. Joint-amplitude mean absolute error
(MAE), in degrees, compares the same six joint ranges using all complete
one-second windows, including the first. We also report a Sinkhorn discrepancy
in common evaluation coordinates. This discrepancy depends partly on the
mixture of early and late histories, which changes with episode duration.
It must therefore be interpreted alongside gait metrics. Contact-speed and left--right imbalance
metrics, exact window definitions, and all milestone tables are provided in
the supplementary results. We report all methods at the final budget and
show their learning curves at the intermediate checkpoints.

\section{Results}
We evaluate motion acquisition, reward generalization beyond the rollout,
the cost of dynamic versus static supervision, and attenuation in barycentric
targets.

\subsection{Final-budget motion acquisition}
\begin{table*}[t]
\centering\small\setlength{\tabcolsep}{4.0pt}
\caption{Final physical comparison at 50M transitions. Speed, excursion, and falls distinguish locomotion quality from balance alone.}
\label{tab:final50m}
\begin{tabular}{llrrrrrrr}
\toprule
Reset & Method & Mean steps & Falls & Speed (m/s) & Excursion & Amp. MAE ($^\circ$) & Sinkhorn & Late $n$\\
\midrule
Dataset & FM potential (ours) & 500.0 & 0 & 0.727 & 1.119 & 9.95 & 0.159 & 64\\
 & Endpoint & 500.0 & 0 & 0.571 & 0.997 & 9.44 & 0.158 & 64\\
 & Value & 500.0 & 0 & 0.776 & 1.089 & 8.69 & 0.186 & 64\\
 & 16-bank score & 500.0 & 0 & 0.518 & 0.891 & 9.25 & 0.196 & 64\\
 & AMP & 498.8 & 3 & 0.478 & 1.035 & 9.37 & 0.239 & 64\\
 & Barycentric OT & 499.3 & 3 & 0.000 & 0.025 & 38.44 & 0.347 & 64\\
\midrule
Fixed & FM potential (ours) & 469.4 & 129 & 0.338 & 0.754 & 12.60 & 0.285 & 61\\
 & Endpoint & 453.8 & 243 & 0.147 & 0.703 & 13.38 & 0.285 & 64\\
 & Value & 451.6 & 315 & 0.047 & 0.387 & 25.23 & 0.308 & 63\\
 & 16-bank score & 466.7 & 196 & -0.021 & 0.667 & 14.27 & 0.526 & 62\\
 & AMP & 500.0 & 0 & 0.000 & 0.027 & 39.24 & 0.424 & 64\\
 & Barycentric OT & 500.0 & 0 & 0.000 & 0.144 & 35.15 & 0.557 & 64\\
\bottomrule
\end{tabular}
\end{table*}

Table~\ref{tab:final50m} compares all methods after 50M transitions.
Under dataset reset, FMP and the three scalar-score controls reach the
500-step episode cap without falls. FMP achieves 0.727\,m/s, compared with
0.478\,m/s for AMP. Value reaches 0.776\,m/s and has lower amplitude error
than FMP. Endpoint has nearly the same Sinkhorn discrepancy as FMP.
Barycentric OT reaches near-cap survival but produces negligible forward
motion and only 0.025 expert-relative excursion.

Fixed-pose initialization separates the methods more clearly. FMP reaches
0.338\,m/s with 0.754 expert-relative excursion and 129 falls in 1,024
episodes. Endpoint reaches 0.147\,m/s with 0.703 excursion and 243 falls.
Both methods have the same rounded Sinkhorn discrepancy (0.285).
Value produces less motion (0.047\,m/s and 0.387 excursion).
The direct 16-bank score has slightly negative net forward speed. AMP and barycentric OT survive all
500 steps but remain nearly stationary, with excursions of 0.027 and 0.144.
Survival alone therefore does not establish that a policy has acquired the
demonstrated gait. Within the learned scalar-reward family, path supervision
improves motion acquisition.
Across all 1,024 first episodes, mean signed forward displacement is 2.97\,m
for FMP and 2.42\,m for Endpoint.

\begin{figure*}[t]
\centering
\includegraphics[width=.99\textwidth]{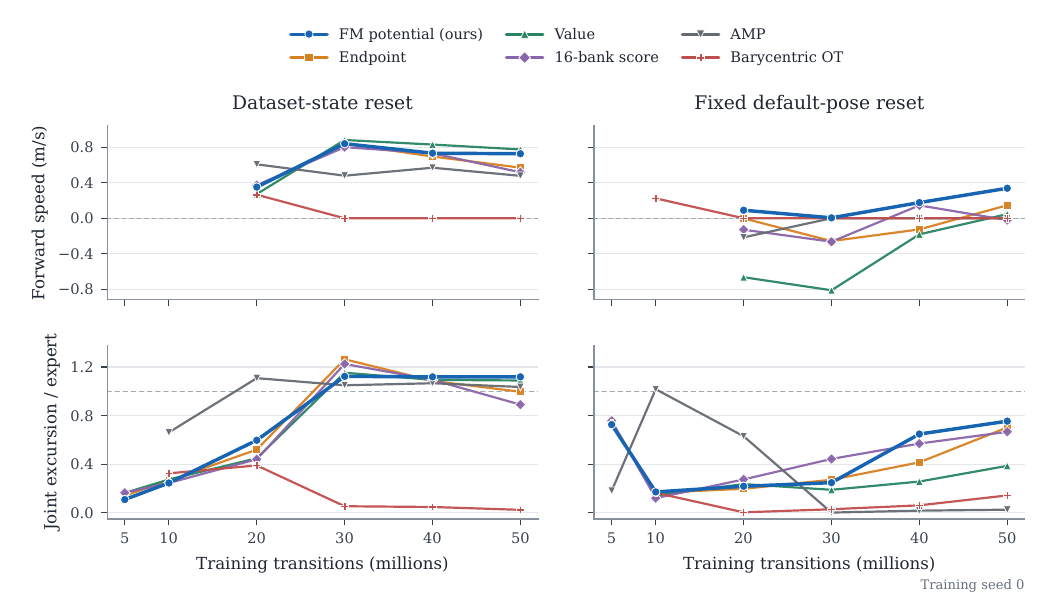}
\caption{Motion acquisition across training budgets under dataset and fixed-pose
initialization. Late speed and expert-relative joint excursion distinguish
walking from stationary survival.}
\label{fig:learning}
\end{figure*}

The learning curves show that forward motion develops after balance
(Fig.~\ref{fig:learning}). Under fixed-pose reset, FMP reaches near-cap
survival early, but its forward speed rises from 0.005\,m/s at 30M to
0.177\,m/s at 40M and 0.338\,m/s at 50M.

\subsection{Reward-level evidence for path supervision}
We first evaluate reward fitting with the policy held fixed. The fitting pool
contains 32,768 histories from a stationary policy. Environments 0--767
provide training queries. The 8,192 histories from environments 768--1023
are held out. All models share the transport pool and are evaluated
against independent 128-bank reference ensembles. After 9,216 updates,
held-out reward MAE is 0.059 for FMP, 0.069 for Endpoint, and 0.094 for
Value. Their gradient cosines are 0.937, 0.923, and 0.382, respectively. Endpoint supervision
improves local gradient agreement, and path supervision further reduces
reward error.

\begin{figure*}[t]
\centering
\includegraphics[width=.99\textwidth]{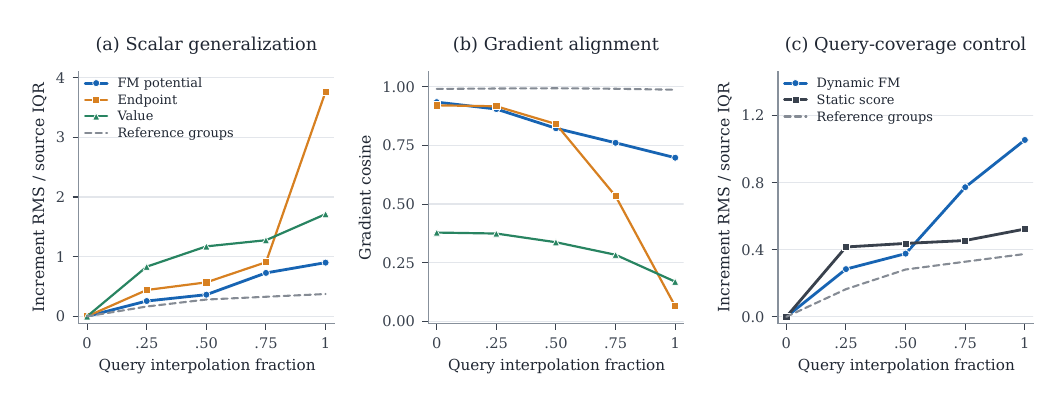}
\caption{Reward geometry on common rollout-to-expert interpolation queries.
In (c), dynamic FM has lower increment error at fractions 0.25 and 0.50,
whereas static supervision has lower error at 0.75 and 1.00.}
\label{fig:mechanism}
\end{figure*}

The larger gain appears beyond the fitting rollout. On common rollout-to-expert
interpolation queries, expert-endpoint score correlation is 0.979 for FMP,
0.157 for Endpoint, and 0.826 for Value (Fig.~\ref{fig:mechanism}a,b).
Independent reference groups correlate at 0.992. All models are evaluated at
flow time zero, so these measurements test the scalar function used for
rewards. Score increments measure the change from each source query to its
interpolated query. Their root mean square errors (RMSE) are normalized by
the source reference IQR. Dashed curves show disagreement between reference groups.

\begin{table}[t]
\centering\small\setlength{\tabcolsep}{4pt}
\caption{Reward generalization on histories executed by stationary and walking
policies. Higher correlation and gradient cosine, and lower normalized RMSE,
indicate closer agreement with the transport reference.}
\label{tab:realqueries}
\begin{tabular}{lrrr}\toprule
Model & Correlation $\uparrow$ & Cosine $\uparrow$ & RMSE/IQR $\downarrow$\\\midrule
Value & 0.618 & 0.264 & 1.473\\
Endpoint & 0.463 & 0.560 & 1.411\\
FM potential & 0.898 & 0.829 & 0.667\\
Reference groups & 0.994 & 0.994 & 0.296\\
\bottomrule\end{tabular}
\end{table}

We also test generalization on physically executed motion. This evaluation
uses 128 stationary and 128 walking histories sampled 0.34--0.64\,s after reset.
The feature coordinates and reference potential remain the same.
In Table~\ref{tab:realqueries}, correlation and gradient cosine are computed
on walking histories. Centered RMSE uses both cohorts and is normalized by
reference-score IQR. FMP achieves the highest correlation and
gradient cosine and the lowest normalized error among the learned models.
Thus the improvement extends to executed histories beyond the stationary
fitting data.

\subsection{Static teacher comparison}
The static-path control tests whether inexpensive pair-displacement labels
provide useful supervision compared with direct score-gradient labels on the
same queries (Fig.~\ref{fig:mechanism}c). At interpolation fractions 0.25 and
0.50, dynamic FM has lower normalized score-increment error: 0.284 and 0.376,
versus 0.416 and 0.437 for static supervision. Static supervision is more
accurate at 0.75 and 1.00. At the expert endpoint, its error is 0.523 versus
1.053 for dynamic FM. Its gradient cosine is also higher: 0.934 versus 0.744.

These fractions specify query location along the rollout-to-expert path.
The network's flow-time input is a separate variable. It is fixed to zero
in both this comparison and online reward evaluation. At source queries,
reward MAE is approximately 0.065 for both models. Walking-query score correlations are
0.880 for dynamic FM and 0.918 for static supervision. The accuracy benefit
is therefore local to the sampled source-side interpolation queries.
Dynamic FM avoids the additional teacher solve required by static supervision.
This reduces offline fitting time from 1,606 to 1,139\,s, a saving of 29\%.

\subsection{Why a naive barycentric target is insufficient}
To examine the motion attenuation introduced by barycentric averaging, we
reconstruct 2,048 first-rollout targets per reset regime. Let $v_j$ denote
an expert sagittal leg-velocity vector and $p_j(h)$ its matching probability
for query $h$. Define norm retention as
\begin{equation}
 \rho=\frac{\E_{h}\|\sum_jp_j(h)v_j\|}
 {\E_{h}\sum_jp_j(h)\|v_j\|}.
 \label{eq:retention}
\end{equation}
Norm retention is 0.130 under dataset reset and 0.103 under fixed reset:
averaging removes approximately 87\% and 90\% of sagittal leg-velocity
magnitude while retaining target forward speeds of 0.851 and 0.834\,m/s.
The matching weights are diffuse, with effective atom count
$N_{\rm eff}(h)=1/\sum_jp_j(h)^2$ averaging 218.4 and 239.7 out of 256.
The reconstructed rewards match the implementation within $2.1\times10^{-7}$.
These measurements identify a key difficulty of barycentric targets:
averaging substantially attenuates joint velocities while retaining a strong
forward-motion target.

\section{Discussion and Limitations}
The central design principle is to learn a scalar preference over realized
motion and to supervise that preference where learning may take the policy.
Transport paths supply queries outside the current rollout. Endpoint
supervision fits the gradient of the score used for rewards, while scalar
calibration fits relative score values on current data. Dynamic FM uses pair
displacements as inexpensive labels for the path queries. In the static
comparison, dynamic FM has lower increment error on sampled source-side
interior queries and takes less time to fit. Static labels are more accurate
near the expert endpoint. Pair displacements thus provide useful supervision
without requiring exact transport gradients at every query. This
interpretation explains why a flow-trained reward model can be useful without
generating a target for the actor.

The conclusions concern the tested reward formulation. The nested ablations
support the benefit of path supervision. They do not show that FM eliminates
conditional averaging in general or that adversarial saturation alone explains
AMP's behavior. Likewise, Sinkhorn discrepancy measures
distributional agreement and must be interpreted alongside physical motion
metrics.

Computational efficiency also depends on the comparison. Dynamic labels
reduce fitting time relative to the static teacher, but scalar calibration
and derivative losses still add reward-model computation. Overall training
cost therefore requires both interaction counts and wall-clock measurements.

Empirical validation is limited to one training seed, one retargeted motion
segment, and simulation. The evaluation episodes characterize variation
within each trained policy. Robustness across training seeds and
generalization to additional motions, perturbations, and hardware remain
untested. Broader validation should also include matched comparisons with
WAIL, PWIL, and SMP.

\section{Conclusion}
We presented an online motion reward learned by flow matching of scalar
transport potentials. The method combines value calibration, endpoint
derivatives, and interior-path supervision. It learns from queries beyond
current policy samples and provides a scalar reward to PPO. Controlled
experiments show improved reward predictions on executed motion histories.
Compared with static supervision, dynamic FM has lower increment error at
source-side interpolation queries and uses 29\% less offline fitting time.
Static labels remain more accurate near the expert endpoint. Under fixed-pose
initialization, path supervision improves aggregate forward progress and
reduces falls relative to endpoint-only fitting. Other scalar controls remain
competitive under demonstration initialization.

\section*{Acknowledgments}
OpenAI Codex was used to assist with language and visualization. All AI-generated
content was reviewed, verified, and edited by the authors, who take full
responsibility for the final content of the manuscript.

\balance
\bibliographystyle{IEEEtran}
\bibliography{references_transport_potential}

@article{peng2021amp,
  title={{AMP}: Adversarial Motion Priors for Stylized Physics-Based Character Control},
  author={Peng, Xue Bin and Ma, Ze and Abbeel, Pieter and Levine, Sergey and Kanazawa, Angjoo},
  journal={ACM Transactions on Graphics},
  volume={40}, number={4}, pages={144:1--144:20}, year={2021},
  doi={10.1145/3450626.3459670}
}

@article{mu2026smp,
  title={{SMP}: Reusable Score-Matching Motion Priors for Physics-Based Character Control},
  author={Mu, Yuxuan and Zhang, Ziyu and Shi, Yi and Yang, Dun and Matsumoto, Minami and Imamura, Kotaro and Tevet, Guy and Guo, Chuan and Taylor, Michael and Shu, Chang and Xi, Pengcheng and Peng, Xue Bin},
  journal={ACM Transactions on Graphics}, volume={45}, number={4},
  pages={1--21}, year={2026}, doi={10.1145/3811282}
}

@inproceedings{dadashi2021pwil,
  title={Primal {Wasserstein} Imitation Learning},
  author={Dadashi, Robert and Hussenot, L{\'e}onard and Geist, Matthieu and Pietquin, Olivier},
  booktitle={International Conference on Learning Representations}, year={2021}
}

@inproceedings{fatras2020minibatch,
  title={Learning with Minibatch {Wasserstein}: Asymptotic and Gradient Properties},
  author={Fatras, Kilian and Zine, Younes and Flamary, R{\'e}mi and Gribonval, Remi and Courty, Nicolas},
  booktitle={International Conference on Artificial Intelligence and Statistics},
  series={Proceedings of Machine Learning Research}, volume={108},
  pages={2131--2141}, year={2020}
}

@inproceedings{feydy2019sinkhorn,
  title={Interpolating between Optimal Transport and {MMD} Using {Sinkhorn} Divergences},
  author={Feydy, Jean and S{\'e}journ{\'e}, Thibault and Vialard, Fran{\c c}ois-Xavier and Amari, Shun-ichi and Trouve, Alain and Peyr{\'e}, Gabriel},
  booktitle={International Conference on Artificial Intelligence and Statistics},
  series={Proceedings of Machine Learning Research}, volume={89},
  pages={2681--2690}, year={2019}
}

@article{liu2022rectifiedot,
  title={Rectified Flow: A Marginal Preserving Approach to Optimal Transport},
  author={Liu, Qiang},
  journal={arXiv preprint arXiv:2209.14577}, year={2022}
}

@inproceedings{balcerak2025energy,
  title={Energy Matching: Unifying Flow Matching and Energy-Based Models for Generative Modeling},
  author={Balcerak, Michal and Amiranashvili, Tamaz and Terpin, Antonio and Shit, Suprosanna and Bogensperger, Lea and Kaltenbach, Sebastian and Koumoutsakos, Petros and Menze, Bjoern},
  booktitle={Advances in Neural Information Processing Systems},
  volume={38}, pages={9514--9540}, year={2025},
  doi={10.52202/085713-0291}
}

@article{xiao2019wail,
  title={{Wasserstein} Adversarial Imitation Learning},
  author={Xiao, Huang and Herman, Michael and Wagner, Joerg and Ziesche, Sebastian and Etesami, Jalal and Linh, Thai Hong},
  journal={arXiv preprint arXiv:1906.08113}, year={2019}
}

@article{barco2026projected,
  title={Projected Energy Matching for Generative {3D} Priors},
  author={Barco, Daniel and Balcerak, Michal and Shit, Suprosanna and Prabhakar, Chinmay and Denzel, Philipp and Menze, Bjoern and Schilling, Frank-Peter},
  journal={arXiv preprint arXiv:2607.07749}, year={2026}
}

@inproceedings{neklyudov2023action,
  title={Action Matching: Learning Stochastic Dynamics from Samples},
  author={Neklyudov, Kirill and Brekelmans, Rob and Severo, Daniel and Makhzani, Alireza},
  booktitle={International Conference on Machine Learning},
  series={Proceedings of Machine Learning Research}, volume={202},
  pages={25858--25889}, year={2023}
}

@article{peng2018deepmimic,
  title={{DeepMimic}: Example-Guided Deep Reinforcement Learning of Physics-Based Character Skills},
  author={Peng, Xue Bin and Abbeel, Pieter and Levine, Sergey and van de Panne, Michiel},
  journal={ACM Transactions on Graphics},
  volume={37}, number={4}, pages={143:1--143:14}, year={2018},
  doi={10.1145/3197517.3201311}
}

@article{peng2022ase,
  title={{ASE}: Large-Scale Reusable Adversarial Skill Embeddings for Physically Simulated Characters},
  author={Peng, Xue Bin and Guo, Yunrong and Halper, Lina and Levine, Sergey and Fidler, Sanja},
  journal={ACM Transactions on Graphics},
  volume={41}, number={4}, note={Art. no. 94}, year={2022},
  doi={10.1145/3528223.3530110}
}

@inproceedings{ho2016gail,
  title={Generative Adversarial Imitation Learning},
  author={Ho, Jonathan and Ermon, Stefano},
  booktitle={Advances in Neural Information Processing Systems},
  volume={29}, year={2016}
}

@article{schulman2017ppo,
  title={Proximal Policy Optimization Algorithms},
  author={Schulman, John and Wolski, Filip and Dhariwal, Prafulla and Radford, Alec and Klimov, Oleg},
  journal={arXiv preprint arXiv:1707.06347}, year={2017}
}

@inproceedings{schulman2016gae,
  title={High-Dimensional Continuous Control Using Generalized Advantage Estimation},
  author={Schulman, John and Moritz, Philipp and Levine, Sergey and Jordan, Michael I. and Abbeel, Pieter},
  booktitle={International Conference on Learning Representations}, year={2016}
}

@inproceedings{arjovsky2017wgan,
  title={{Wasserstein} Generative Adversarial Networks},
  author={Arjovsky, Martin and Chintala, Soumith and Bottou, L{\'e}on},
  booktitle={International Conference on Machine Learning},
  series={Proceedings of Machine Learning Research}, volume={70},
  pages={214--223}, year={2017}
}

@inproceedings{cuturi2013sinkhorn,
  title={{Sinkhorn} Distances: Lightspeed Computation of Optimal Transport},
  author={Cuturi, Marco},
  booktitle={Advances in Neural Information Processing Systems},
  volume={26}, year={2013}
}

@article{peyre2019computational,
  title={Computational Optimal Transport with Applications to Data Sciences},
  author={Peyr{\'e}, Gabriel and Cuturi, Marco},
  journal={Foundations and Trends in Machine Learning},
  volume={11}, number={5--6}, pages={355--607}, year={2019},
  doi={10.1561/2200000073}
}

@article{hyvarinen2005score,
  title={Estimation of Non-Normalized Statistical Models by Score Matching},
  author={Hyv{\"a}rinen, Aapo},
  journal={Journal of Machine Learning Research},
  volume={6}, number={24}, pages={695--709}, year={2005}
}

@inproceedings{lipman2023flow,
  title={Flow Matching for Generative Modeling},
  author={Lipman, Yaron and Chen, Ricky T. Q. and Ben-Hamu, Heli and Nickel, Maximilian and Le, Matt},
  booktitle={International Conference on Learning Representations}, year={2023}
}

@article{tong2024cfm,
  title={Improving and Generalizing Flow-Based Generative Models with Minibatch Optimal Transport},
  author={Tong, Alexander and Fatras, Kilian and Malkin, Nikolay and Huguet, Guillaume and Zhang, Yanlei and Rector-Brooks, Jarrid and Wolf, Guy and Bengio, Yoshua},
  journal={Transactions on Machine Learning Research}, year={2024}
}

@inproceedings{liu2023rectified,
  title={Flow Straight and Fast: Learning to Generate and Transfer Data with Rectified Flow},
  author={Liu, Xingchao and Gong, Chengyue and Liu, Qiang},
  booktitle={International Conference on Learning Representations}, year={2023}
}

@article{harvey2020robust,
  title={Robust Motion In-Betweening},
  author={Harvey, F{\'e}lix G. and Yurick, Mike and Nowrouzezahrai, Derek and Pal, Christopher},
  journal={ACM Transactions on Graphics},
  volume={39}, number={4}, pages={60:1--60:12}, year={2020},
  doi={10.1145/3386569.3392480}
}

@article{mittal2025isaaclab,
  title={{Isaac Lab}: A {GPU}-Accelerated Simulation Framework for Multi-Modal Robot Learning},
  author={Mittal, Mayank and Roth, Pascal and Tigue, James and Richard, Antoine and Zhang, Octi and Du, Peter and others},
  journal={arXiv preprint arXiv:2511.04831}, year={2025}
}

@article{peng2025mimickit,
  title={{MimicKit}: A Reinforcement Learning Framework for Motion Imitation and Control},
  author={Peng, Xue Bin},
  journal={arXiv preprint arXiv:2510.13794}, year={2025}
}
\end{document}